%% file: main.tex
\documentclass[10pt,twocolumn,letterpaper]{article}

\usepackage{iccv}
\usepackage{times}
\usepackage{epsfig}
\usepackage{graphicx}
\usepackage{amsmath}
\usepackage{amssymb}
\usepackage{blindtext}
\usepackage{graphicx}
\usepackage{algorithm2e}
\RestyleAlgo{ruled}
\usepackage{fullpage}
\usepackage{mathtools}
\usepackage{latexsym,amsmath,amssymb}    
\usepackage{caption}
\usepackage[font={small}]{caption}
\usepackage{newfloat}
\usepackage{url}
\usepackage{listings}
\usepackage{enumitem}
\usepackage{subcaption}
\usepackage{todonotes}
\usepackage{booktabs}[demo]
\usepackage{float}
\usepackage{amsmath}
\usepackage{adjustbox}
\usepackage{footnote}
\input{preamble/math}

\usepackage[pagebackref,breaklinks,colorlinks]{hyperref}

\iccvfinalcopy 

\ificcvfinal
\begin{document}

\title{Softmax Bias Correction for Quantized Generative Models}

\author{Nilesh Prasad Pandey, Marios Fournarakis, Chirag Patel, Markus Nagel\\
Qualcomm AI Research\thanks{Qualcomm AI Research is an initiative of Qualcomm Technologies, Inc}\\
{\tt\small \{nileshpr, mfournar, cpatel, markusn\}@qti.qualcomm.com}}


\maketitle

\begin{abstract}
Post-training quantization (PTQ) is the go-to compression technique for large generative models, such as stable diffusion or large language models. PTQ methods commonly keep the softmax activation in higher precision as it has been shown to be very sensitive to quantization noise. However, this can lead to a significant runtime and power overhead during inference on resource-constraint edge devices. In this work, we investigate the source of the softmax sensitivity to quantization and show that the quantization operation leads to a large bias in the softmax output, causing accuracy degradation. To overcome this issue, we propose an offline bias correction technique that improves the quantizability of softmax without additional compute during deployment, as it can be readily absorbed into the quantization parameters. We demonstrate the effectiveness of our method on stable diffusion v1.5 and 125M-size OPT language model, achieving significant accuracy improvement for 8-bit quantized softmax. 

\end{abstract}

\input{sections/01_introduction}\label{sec:introduction}
\input{sections/02_background}\label{sec:background}
\input{sections/03_method}\label{sec:method}
\input{sections/04_experiments}\label{sec:experiments}
\input{sections/05_conclusion}\label{sec:conclusion}
\bibliographystyle{plain}
\bibliography{ref}
\end{document}

%% file: preamble/math.tex
\newcommand\mat[1]{\mathbf{#1}}

\renewcommand\vec[1]{\mathbf{#1}}

\renewcommand\vec[1]{\mathbf{#1}}

\newcommand{\softmax}[1]{\text{softmax}\left(#1\right)}

\newcommand\eop[1]{\mathop{\mathbb{E}}\left[#1\right]}
\newcommand\eopX[2]{\mathop{\mathbb{E}_{#1}}\left[#2\right]}

\newcommand\INT[1]{#1_{\text{int}}}



%% file: sections/01_introduction.tex
\section{Introduction}
The increasing prevalence of large generative neural networks, such as stable diffusion~\cite{dhariwal2021diffusion,ho2020denoising,rombach2022high}, ChatGPT, and OPT~\cite{zhang2022opt}, has revolutionized the fields of computer vision and natural language processing. These models exhibit exceptional capabilities in generating realistic images and human-like text. However, deploying them on edge devices is challenging due to their size and computational demands. To address this issue, quantization has emerged as the most promising technique to optimize model deployment on resource-constrained devices, with a plethora of work emerging for both vision~\cite{Qualcomm,chen2023speed,li2023snapfusion} and language models~\cite{liu2023noisyquant, frantar2023optq, frantar2022OBS}. 

Post-training quantization~(PTQ) is the go-to method for quantizing such models because accessing original training data and pipelines can be difficult, and training them requires vast computing resources. However, activation quantization remains challenging because certain layers, such as the softmax in transformers, are particularly sensitive to quantization. This issue is even more pronounced in diffusion models due to the iterative nature of the denoising process leading to error accumulation. For this reason, it is common practice to keep the softmax unquantized or in higher precision leading to significant latency overhead, especially in networks with larger sequence lengths~\cite{stevens2021softermax}.

In this work, we systematically investigate the source of the softmax sensitivity to quantization and show that quantization operation leads to a large bias degrading accuracy. We introduce a hardware-friendly bias correction that acts as an offset at the softmax output, which can be absorbed into the quantization parameters. Despite its simplicity, our method significantly improves the SQNR and perplexity scores for diffusion~\cite{dhariwal2021diffusion} and OPT~\cite{zhang2022opt} language models, respectively, with 8-bit quantized softmax.

%% file: sections/02_background.tex
\section{Background}
\subsection{Related Work}
\label{sec:related_work}
Quantization is one of the most effective methods available for reducing latency and power consumption in neural network inference. This is achieved not only thanks to reduced model size but because fixed-point operations are more efficient than their floating-point counterparts.  In this work, we focus on post-training quantization (PTQ), which takes a pre-trained FP32
network and converts it directly into a fixed-point network without the need for the original training pipeline~\cite{nagel2020up, nagel2019data, krishnamoorthi2018quantizing}. These methods require either no data or only a small calibration dataset and are easier to use compared to quantization-aware training (QAT)~\cite{esser2019learned, 2021HanBinReg, 2019GongDSQ, Nagel2022Oscillations}. For more details on neural network quantization, we refer the reader to~\cite{Gholami2021QuantSurvey, nagel2021white}.

As the success of language models has increased concurrently with their size, a lot of recent work has focused on quantizing these models~\cite{shen2020QBERT, zafrir2019Q8Bert, shen2020QBERT, frantar2023gptq}. While few methods have emerged to address the issue of outliers in the output of transformers~\cite{bondarenko-etal-2021-understanding, bondarenko2023quantizable, wei2023OutlierSuppression}, our work is complementary as we focus on quantizing the attention weights. 

Similarly, while recent work on the quantizing diffusion models \cite{li2023QDiffusion,he2023ptqd,shang2023PTQDiffusion} have discussed various problems and methods to overcome quantization challenges, most of these methods keep sensitive activations, such as softmax, in higher-precision. However, softmax can be the biggest latency bottleneck due to its inefficient execution in hardware~\cite{stevens2021softermax}. Our work is orthogonal to existing methods as we focus on improving the quantizibility of softmax layers to lower bits.

\subsection{Motivation}
\label{sec:motivation}
Softmax accounts for a significant fraction of the total runtime of transformers accounting for up to 40\% for sequence lengths larger than 2048~\cite{stevens2021softermax}. As a result, keeping softmax in low precision can accelerate inference by reducing the size of the look-up tables required to estimate exponential functions. As modern diffusion model, \eg stable diffusion v1.5~\footnote{\url{https://github.com/runwayml/stable-diffusion}} reach sequence lengths of 4096, low-bit softmax is imperative if we want to achieve competitive on-device performance. 

However, when quantizing the softmax in stable diffusion to 8 bits, we observe a considerable deviation in the generated images compared to the floating-point model (see columns FP32 and W8A16-SM8 in figure~\ref{fig:comparison}). On the contrary, when keeping softmax at 16 bits (column W8A16 in figure~\ref{fig:comparison}), the generated image matches that of the floating-point model very closely. 

To confirm our hypothesis that the softmax layers in the diffusion process are particularly sensitive to quantization, we perform the following sensitivity analysis: we quantize individual attention tensors to 8 bits in the denoising U-Net while keeping the rest of the network in FP32 and measure the signal-to-quantization noise ratio (SQNR) between the quantized and full-precision at the end of the denoising process. We use a calibration set  $\mathcal{X}$  of 400 input latents sampled uniformly across all time steps and report the mean SQNR in dB in table~\ref{tab:attn_sensitivity_analysis}. We calculate the SQNR using the following formula:
\begin{equation}\label{eq:SQNR} 
    \text{SQNR}_{\text{dB}}= 10\log{
    \eopX{\vec{x}}{\frac{\|\phi\left(\vec{x}\right)\|_2^2}{\| q\left(\phi\left(\vec{x}\right)\right) -\phi\left(\vec{x}\right)  \|_2^2}}},
\end{equation}
where $\vec{x} \in \mathcal{X}$, $\phi(\cdot)$ is the output of the denoising U-Net,  and $\| \cdot \|_2$ is the Forbenious norm. 

We can see from table~\ref{tab:attn_sensitivity_analysis} that quantizing the softmax output leads to an 8-fold degradation in SQNR compared to the second most sensitive activation, the value tensor~(V).  

\begin{table}[t]
    \centering
    \fontsize{9.25pt}{9.25pt}\selectfont
    \begin{tabular} {  lcccc }
    \toprule
    Activation in 8 bits & SQNR($\uparrow$) \\
    \midrule
    Query~(Q) & $32.36$ \\
    Key~(K) & $29.77$ \\
    Value~(V) & $26.58$\\
    Attention score (softmax input)
& $28.09$ \\
    Softmax output & $3.24$ \\
    \bottomrule
\end{tabular}
\caption{Quantization sensitivity analysis for attention layers in the denoising U-Net of stable diffusion. We quantize each activation to 8 bits while keeping the rest of the network unquantized, and report mean SQNR($\uparrow$): the higher, the better.} 
    \label{tab:attn_sensitivity_analysis}
\end{table}
\subsection{Quantized softmax is biased}
\begin{figure}[t]
    \centering
    \includegraphics[width=0.35\textwidth]{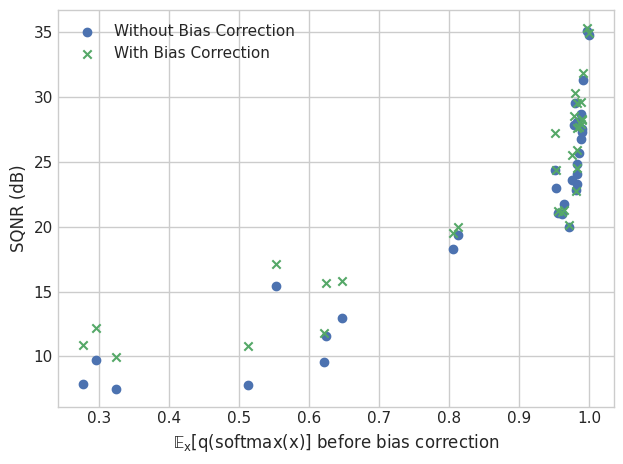}
    \caption{x-axis: sum of 8-bit quantized softmax vectors before bias correction while keeping rest of the network in FP32 $\eopX{\vec{x}}{q(\softmax{\vec{x}}})$; y-axis:  SQNR between full-precision and quantized UNet outputs after final diffusion step.}
    \label{fig:expectedsum}
\end{figure}

Why is the softmax output so sensitive to quantization? Having a closer look at the values of the quantized softmax, we found that up to 99\% of the values are rounded to zero. 
As all these values are rounded down, the resulting quantization error is \emph{biased}, and the softmax probabilities are not correctly normalized anymore, which can degrade the model's performance.
In the scatter plot of figure~\ref{fig:expectedsum}, we see that many quantized  softmax outputs do not add up to 1.0. In fact, the expected sum of the softmax output over the calibration set can be as low as 0.3. We also observe a high correlation between quantization bias and degradation of the denoising process: the larger gap from the expected softmax output (1.0), the lower the SQNR at the U-Net output.  

%% file: sections/03_method.tex
\section{Quantized activation bias correction}
In the previous section, we experimentally established that quantizing the softmax in the transformer can lead to highly-biased outputs causing significant distortion in stable diffusion's output. In this section, we outline a simple but effective method for correcting this bias and improving performance. 

We define quantization bias as the systematic discrepancy between quantized and unquantized activation vector $\vec{y}$: 
\begin{equation}
    \beta\left(\vec{y};\mathcal{T} \right) = \eop{\mathcal{T}\vec{y}} - \eop{q\left(\mathcal{T}\vec{y}\right)}, 
    \label{eq:quantization_bias_definition}
\end{equation}
where $\mathcal{T}$ is the transformation function acting on $\vec{y}$, and $q(\cdot)$ is the quantization function. The transformation $\mathcal{T}$ could be the identity or a simple linear transformation, such as a reduction along a certain axis. We can now \textit{correct} for this bias by adding back to the quantized  activation $\vec{y}_q= q\left(\mathcal{T}\vec{y}\right)$, such that
\begin{align}
\eop{\mathcal{T}\vec{y}_q + \beta\left(\vec{y};\mathcal{T} \right)} & = \eop{\mathcal{T}\vec{y}}. 
\end{align}
In practice, we calculate an empirical estimate of the bias, $\widehat{\beta}$, using the available calibration data.  

\subsection{Softmax bias correction}
\label{sec:softamx_bias_correction}
In the case of softmax activations, we know in advance that its output is normalized and should thus sum to 1.0. Using the notation of equation~\eqref{eq:quantization_bias_definition}, the transformation $\mathcal{T}$ is an inner product with the vector of ones along the normalization dimension: $\eop{\vec{1}^{\top}\vec{y}}=1$. In transformers, the input to the softmax layer is typically three-dimensional $\mat{X}\in \mathbb{R}^{n_{\text{heads}} \times n_{\text{seq}} \times n_{\text{seq}}}$ and the softmax is applied across the last dimension. Depending on the capabilities of the hardware available, we could have a \textit{per-tensor} or \textit{per attention-head} correction factor, which would require reducing the output of the softmax output accordingly. For example, the \textit{per-tensor} correction factor is calculated by:
\begin{equation}
    \beta = \frac{1}{n_{\text{seq}}} - \frac{\eopX{\vec{x}}{\sum_{i}^{n_{\text{heads}}} \sum_{j}^{n_{\text{seq}}} \sum_{k}^{n_{\text{seq}}}{\mat{Y}_{i,j,k}}}}{n_{\text{heads}}  n_{\text{seq}}^2}, 
    \label{eq:pt_bias_correction}
\end{equation} 
where $\mat{Y}=q\left(\softmax{\mat{X}} \right)$ and $\beta$ is added elementwise to the whole quantized output $\mat{Y}$. In later sections~(cf. Sec~\ref{sec:bias_corr_granularity}), we perform an ablation study for bias correction granularities. 

\subsection{Absorbing bias correction}
\label{sec:absorbing_bias_correction}
An important benefit of our bias correction method is that it can be easily absorbed into the offset of \textit{asymmetric} quantization. For a $b$ bitwidth uniform quantizer with scale $s$ and zero-point $z$, asymmetric quantization is defined as: 
\begin{equation}
    \begin{aligned}
    \centering
    y_q & = q(y;s,z,b) \\
    &= s\cdot \left[ \mathrm{clamp} \left(\left\lfloor \frac{y}{s}\right\rceil + z,0,2^b-1\right)-z \right] \\
    & =  s \cdot \INT{y}- c ,
    \end{aligned}
    \label{eq:asymmetric_quant_def}
\end{equation}
where $c= s \cdot z$ is a floating-point offset of the quantization grid, as the scale $s$ is typically a floating-point number~\cite{krishnamoorthi2018quantizing}. For bias correction, we only have to absorb the correction factor into the offset, $c' =  s \cdot z - \beta$, while keeping everything else the same. Given that activations are most commonly asymmetrically quantized~\cite{bondarenko-etal-2021-understanding, wei2023OutlierSuppression, nagel2021white, bondarenko2023quantizable}, our bias correction leads to no additional compute.

%% file: sections/04_experiments.tex
\section{Experiments}
In this section, we demonstrate the advantages of softmax bias correction. We perform experiments on stable diffusion v1.5\footnote[2]{\url{https://github.com/huggingface/diffusers}} and extend our analysis to a transformer-based language model. We experiment with the 125M-sized variant of OPT~\cite{zhang2022opt} pre-trained using the causal language modeling (CLM) objective. We use a validation pipeline for HuggingFace libraries~\cite{Wolf2020SOTATransformers, lhoest2021datasets} and evaluate on Wikipedia validation set\footnote{Specifically, we use the English subset of Wiki-40b, \url{https://huggingface.co/datasets/wiki40b}, that contains cleaned-up text of English Wikipedia and training/validation splits.}. We report SQNR between the full-precision and quantized U-Net output for stable diffusion~(the higher, the better), and CLM perplexity for OPT~(the lower, the better).

We use PyTorch v1.11 and the AI Model Efficiency Toolkit (AIMET)\footnote[3]{AIMET is a product of Qualcomm Innovation Center, Inc., available on GitHub at \url{https://github.com/quic/aimet}}  \cite{siddegowda2022neural} to quantize the models to desired bitwidths. We implement per-tensor symmetric quantization for weights and asymmetric quantization for activations.


\begin{figure*}[t]
    \centering
    \subfloat[\centering First shot of the Milky Way
]{\includegraphics[width=0.5\textwidth]{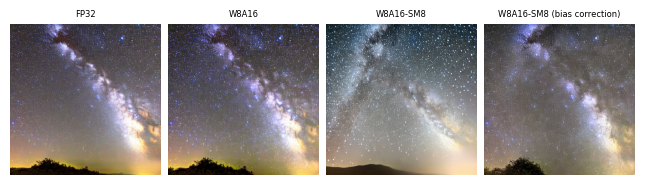}}
     \subfloat[\centering Tuscany, Italy, Country Side, Sunrise, Foggy, Dawn, Landscape, Aerial view, Meadow, 5K]{\includegraphics[width=0.5\textwidth]{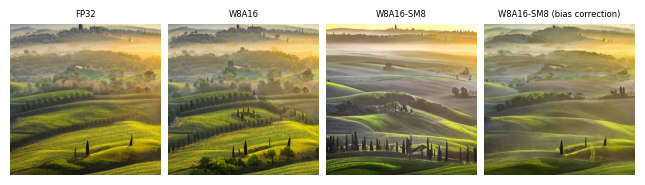}} 
     \\
     \subfloat[\centering skeleton in heaven dressed as renaissance artist painting a portrait 
     of a model dressed as a Saint, painting, renaissance art, detailed,oil painting]{\includegraphics[width=0.5\textwidth]{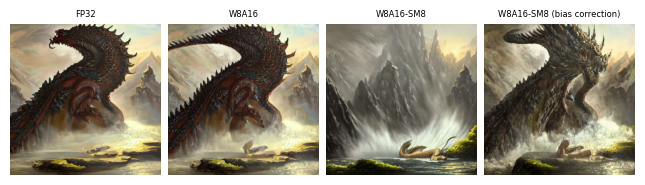}} 
     \subfloat[\centering highly detailed oil painting of a western dragon emerging from a hot spring, fantasy, featured on art station
]{\includegraphics[width=0.5\textwidth]{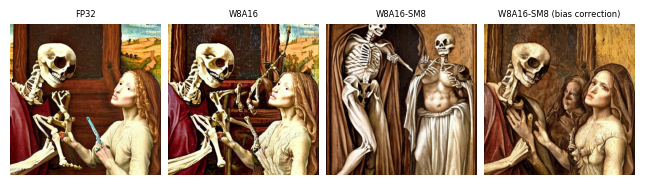}}
    \caption{\protect{Visual comparison between FP32, W8A16, W8A16 with softmax quantized to 8 bits without and with per attention head based bias correction generated using 20 diffusion steps on test prompts from LAION Aesthetics dataset \cite{schuhmann2022laion}.} }
    \label{fig:comparison}
\end{figure*}

\begin{table}[t]
    \addtolength{\tabcolsep}{-2pt}
    \centering
    \fontsize{9.25pt}{9.25pt}\selectfont

    \begin{tabular} {  lccc }
    \toprule
    \centering Type of correction & SD (SQNR$\uparrow$) \\
    \midrule
    None  & $3.17$ \\
    Per-tensor  & $5.77$ \\
    Per attention-head & $6.05$\\
    Time-step aware, per-tensor  & $5.93$ \\
    Time-step aware, per attention-head & $6.06$\\

    \bottomrule
\end{tabular}
    \caption{Granularity ablation study for bias correction: we quantize the softmax to 8 bits, keeping the rest of the network in FP32. We report SQNR for stable diffusion (SD) ($\uparrow$): the higher, the better} 
    \label{tab:performance_vs_time step}
\end{table}

\subsection{Granularity of bias correction}
\label{sec:bias_corr_granularity}
As mentioned in section~\ref{sec:softamx_bias_correction}, depending on the target hardware, we can apply bias correction at different granularities of attention tensor, \eg \textit{per-attention head} or \textit{per-tensor}. Due to the iterative nature of the denoising process in diffusion models, activation distributions in the U-Net are time-step dependent, hence motivating us to use the time-step as an additional axis of granularity to perform \textit{time-step-aware} bias correction. We compare the results from the different schemes in table \ref{tab:performance_vs_time step}.

We observe that the per-attention head correction scheme performs on par or better than all other schemes, making it a favorable choice for on-device deployment due to its minimal computational overhead compared to its time-aware counterpart.

\subsection{Main Results}
\label{sec:main_results}
We extend our analysis to include the 125M-size OPT language model, and we report results using per-attention head bias correction in table \ref{tab:performance_mainresults}. We quantize softmax to 8 bits~(SM8) and keep the rest of the network at either full-precision~(FP32) or 8-bit weights and 16-bit activations~(W8A16). With bias correction, we achieve over 2.7dB improvement for stable diffusion and roughly 4.8 improvement in perplexity for OPT in both quantization settings. In the case of diffusion, we also demonstrate the improvement visually in figure \ref{fig:comparison}, by showing the generated images of the quantized diffusion with and without bias correction. As we can see, the generated image with bias correction very closely resembles the full precision output. 

\begin{table}[t]
    \addtolength{\tabcolsep}{-4pt}
    \centering
    \fontsize{9.25pt}{9.25pt}\selectfont

    \begin{tabular} {  lccc }
    \toprule  Configuration & SD (SQNR$\uparrow$) & OPT (ppl$\downarrow$) \\
    \midrule
    \midrule
    FP32 baseline &  -  & 27.73 \\
    \midrule
    FP32-SM8 & 3.17 & 34.98 \\
    FP32-SM8 + bias correction & 6.05 & 30.19 \\
    \midrule 
    \midrule 
    W8A16 baseline & 9.66 & 27.77  \\
    \midrule
    W8A16-SM8 & 3.05 & 35.11 \\
    W8A16-SM8 + bias correction & 5.76 &  30.24 \\
    \bottomrule
\end{tabular}
    \caption{Per-attention bias correction for stable diffusion~(SD) and 125M OPT model for different quantization configurations (W8A16 \& FP32), while keeping the softmax to 8 bits (SM8). We report SQNR for stable diffusion (SD) and CLM perplexity for OPT. ($\uparrow$): the higher, the better; ($\downarrow$): the lower, the better.} 
    \label{tab:performance_mainresults}
\end{table}

%% file: sections/05_conclusion.tex
\section{Conclusions}

In this work, we investigated a common prevailing issue of softmax sensitivity to quantization in the case of generative models. To understand the source of the softmax sensitivity to quantization, we analyzed the softmax distributions and showed that the quantization operation led to a significant bias in the softmax output degrading the performance of generative models when represented in lower precision. To overcome this issue, we proposed a simple yet effective hardware-friendly offset correction to improve the quantizability of softmax layers to lower bits, which is key to achieving competitive on-device performance, especially for transformer-based networks with longer sequence lengths. We demonstrated the effectiveness of our method on stable diffusion v1.5 and 125M-size OPT language model, achieving over 2.7dB improvement for stable diffusion and roughly 4.8 improvement in perplexity for OPT respectively for the 8-bit weights and 16-bit activations (W8A16) quantization setting.
